\documentclass[letterpaper]{article} 
\usepackage{conference} 
\usepackage{times}
\usepackage{helvet}
\usepackage{courier}
\usepackage[hyphens]{url}
\usepackage{graphicx}
\usepackage{natbib}
\usepackage{caption}
\usepackage[ruled,linesnumbered]{algorithm2e}

\usepackage{booktabs}
\usepackage{multirow}
\usepackage{subfigure}
\usepackage{amsmath}
\usepackage{amssymb}
\usepackage{amsfonts}
\usepackage{adjustbox}
\def\ie{{\em i.e.}}
\def\eg{{\em e.g.}}

\nocopyright

\title{Patched Line Segment Learning for Vector Road Mapping}
\author{
    Jiakun Xu\textsuperscript{\rm 1},
    Bowen Xu\textsuperscript{\rm 1},
    Gui-Song Xia\textsuperscript{\rm 1},
    Liang Dong\textsuperscript{\rm 2},
    Nan Xue\textsuperscript{\rm 3}
}
\affiliations{
    \textsuperscript{\rm 1}School of Computer Science, Wuhan University\quad
    \textsuperscript{\rm 2}Google Inc. \quad
    \textsuperscript{\rm 3}Ant Group\\
    
    \{jiakun.xu,\ xbw,\ guisong.xia\}@whu.edu.cn, 
    lidong@google.com,
    xuenan@ieee.org
}

\begin{document}

\maketitle

\begin{abstract}
This paper presents a novel approach to computing vector road maps from satellite remotely sensed images, building upon a well-defined Patched Line Segment (PaLiS) representation for road graphs that holds geometric significance. Unlike prevailing methods that derive road vector representations from satellite images using binary masks or keypoints, our method employs line segments. These segments not only convey road locations but also capture their orientations, making them a robust choice for representation. More precisely, given an input image, we divide it into non-overlapping patches and predict a suitable line segment within each patch. This strategy enables us to capture spatial and structural cues from these patch-based line segments, simplifying the process of constructing the road network graph without the necessity of additional neural networks for connectivity. In our experiments, we demonstrate how an effective representation of a road graph significantly enhances the performance of vector road mapping on established benchmarks, without requiring extensive modifications to the neural network architecture. Furthermore, our method achieves state-of-the-art performance with just 6 GPU hours of training, leading to a substantial 32-fold reduction in training costs in terms of GPU hours.
\end{abstract}

\section{Introduction}\label{sec:intro}

%

\begin{figure}[htb]
\centering
\subfigure[\footnotesize Input image]{
    \begin{minipage}[t]{0.31\linewidth}
    \centering
    \includegraphics[width=\linewidth]{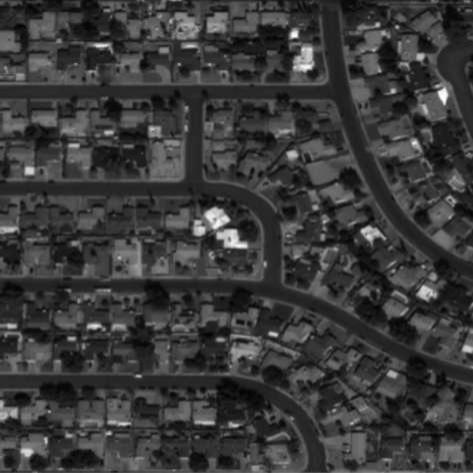}
    \end{minipage}
    \label{fig_teaser_image}
}
\hspace{-4mm}
\subfigure[\footnotesize Ground truth]{
    \begin{minipage}[t]{0.31\linewidth}
    \centering
    \includegraphics[width=\linewidth]{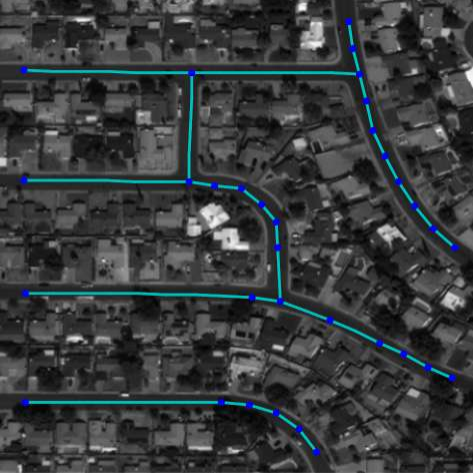}
    \end{minipage}
    \label{fig_teaser_raw_label}
}
\hspace{-3mm}
\subfigure[\footnotesize Dense GT]{
    \begin{minipage}[t]{0.31\linewidth}
    \centering
    \includegraphics[width=\linewidth]{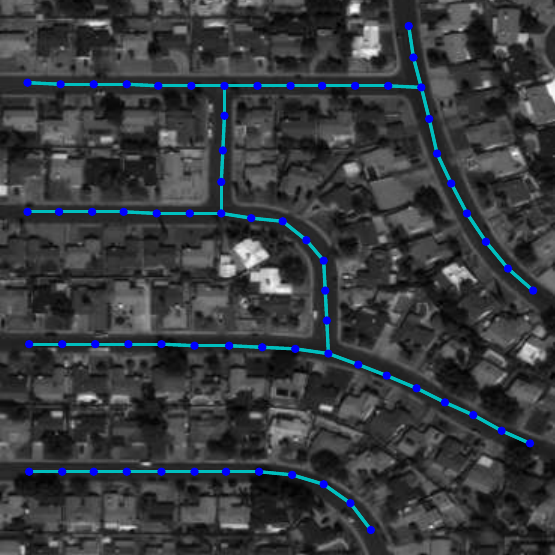}
    \end{minipage}
    \label{fig_teaser_dense_label}
}
\subfigure[\footnotesize Road graph learned from Keypoints~\cite{xu2023rngdet++}]{
    \begin{minipage}[t]{0.95\linewidth}
    \centering
    \includegraphics[width=\linewidth]{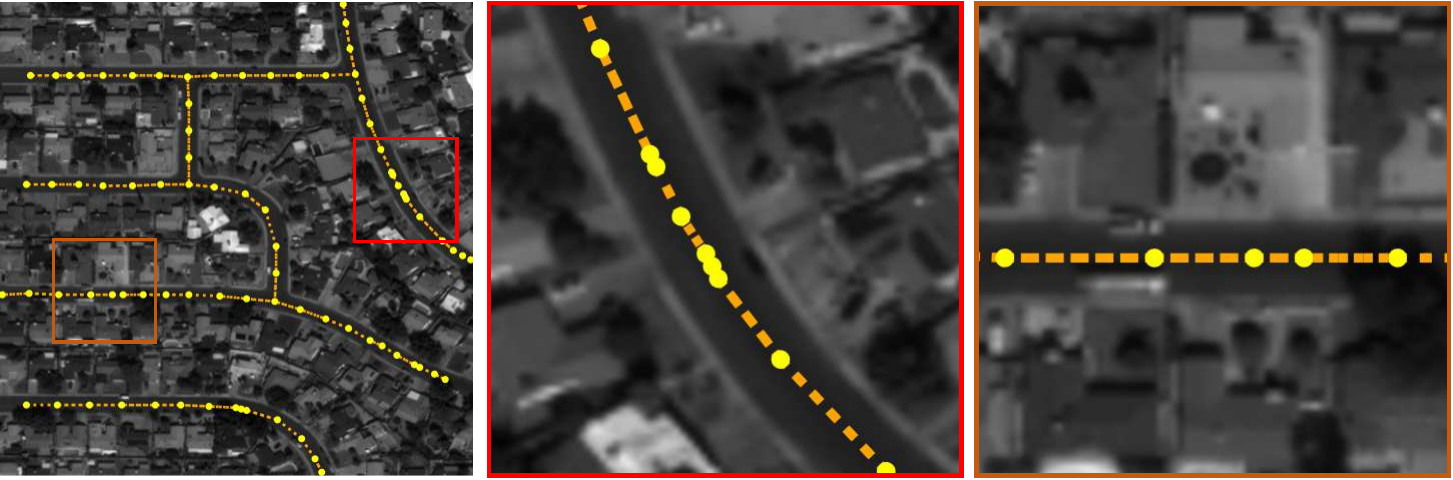}
    \end{minipage}
    \label{fig_teaser_keypoints}
}
\subfigure[\footnotesize Road graph learned via PaLiS (Ours)]{
    \begin{minipage}[t]{0.95\linewidth}
    \centering
    \includegraphics[width=\linewidth]{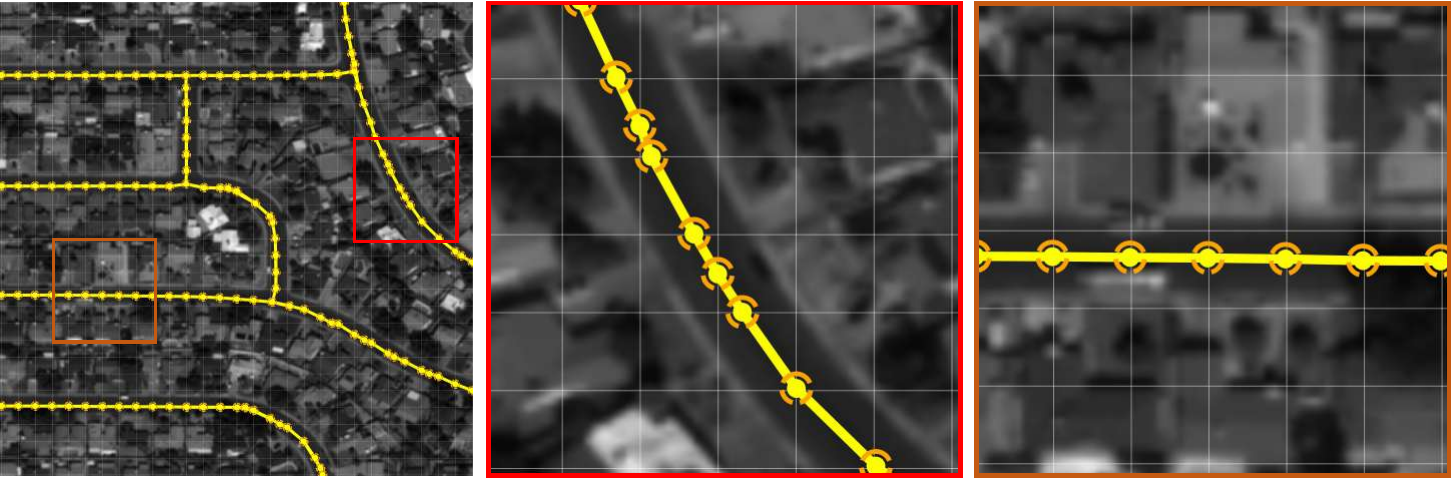}
    \end{minipage}
    \label{fig_teaser_linesegments}
}
\caption{Illustration of graphs constructed by different representations. The predicted representations~(keypoints and line segments) are denoted in yellow marks and the connectivities are denoted in orange marks.}
\label{fig_teaser}
\end{figure}

By ``vector road mapping'', it refers to a process of converting the road features presented in satellite-borne remote sensing images into vector-based and symbolic graph representations, which is also known as {\em road graph extraction} or {\em road network extraction} within the community of remote sensing and plays a fundamental role in numerous downstream tasks including navigation~\citep{zhang2021cycling,cai2023dynamic}, urban planning~\citep{shi2019urban,xu2023hisup}, and autonomous driving~\citep{liebenwein2020compositional,xu2021icurb,9706674,buchner2023learning}. 

The state-of-the-art methods for vector road mapping primarily rely on the strong representation capabilities of deep neural networks. These approaches formulate the problem as a supervised learning task, utilizing paired satellite images and annotated road graphs that use vertices and edges to depict the line and curve structures of roads. As the input images are in pixel form, it becomes crucial to establish an appropriate representation for facilitating the learning from the pixels of satellite images to the vector representation of roads. In the state-of-the-art methods~\cite{batra2019improved,he2020sat2graph,xu2023rngdet++}, the ``appropriate representation'' of vector road annotations were initially come down to mask-based representation (\ie, road masks) and were then upgraded to the keypoint-based graph representation as the main representation in the pursuit of end-to-end learning. 

While keypoint-based graph representations have demonstrated remarkable performance, many of these methods encounter a significant drawback: the substantial training cost involved. For instance, RNGDet++ model~\cite{xu2023rngdet++} requires approximately $192$ GPU hours to train on a dataset of moderate size with thousands of images. This high training cost can be attributed to the prevalent oversampling strategy used to define the ``keypoints'' in the original annotations (depicted in Fig.~\ref{fig_teaser_raw_label}). This strategy involves densely sampling numerous points along each road, as shown in Fig.~\ref{fig_teaser_dense_label}, lacking invariance to commonly employed image transformations used for data augmentation, such as random cropping and image translation, and eventually results in ambiguity during the learning process.
As a consequence, methods employing keypoint-based graph representations must grapple with inherent representation ambiguity, requiring a greater number of training iterations. Such a prolonged training process often entails cluttered patterns in the keypoint detection outcomes, as illustrated by the enclosed regions in Fig.~\ref{fig_teaser_keypoints}. Furthermore, the keypoint-based representations have to leverage additional modules to learn the connectives for the learned keypoints on the fly to accomplish the task of vector road mapping.

In this paper, we devote ourselves to finding a better representation of vector road annotations, to eliminate the ambiguity in the existing keypoint-based graph representations, for the sake of efficient learning during training and top-performing mapping results in the testing phase. Our study is motivated by the recently-proposed PaRK-Detect~\cite{xie2023park} that defines {\em patched keypoints}, in which each small patch (\eg, $16\times 16$) will have at most one keypoint for learning. Because the local patches are uniformly distributed over the image grids, such a definition largely eliminates the ambiguity for learning. However, since the keypoints are unary primitives that did not explicitly define the spatial relationships, PaRK-Detect~\cite{xie2023park} only obtained comparable performance in testing. Motivated by this, we are interested in presenting a patched representation to take its ambiguity-free merits while retaining the spatial context for facilitating the final vector road mapping. 

Our work is inspired by an observation that {\em the spatial and geometric information of roads in local patches can be well represented by line segments instead of keypoints}. Based on this, we present a novel PaLiS (Patched Line Segment) representation to depict the annotated road graphs in a geometrically-meaningful way while enjoying the ambiguity-free merits of patch-based representation. 
By dividing the grid of input images into a set of local (\eg, $8\times 8$) patches, most of the local patches that contain a fragment of road path can uniquely define the only local line segment. To preserve the rich structural information of the local line segments, we use the closed-form $xy-xy$ representation for the two endpoints of a line segment, which facilitates the computation of patch adjacency in a geometrically-meaningful way. As shown in Fig.~\ref{fig_teaser_linesegments}, our proposed PaLiS representation could handle a variety of road graph patterns in a unified representation. With the proposed PaLiS representation, we find out that our PaLiS representation can be reliably learned via the rasterized road masks as supervision in differentiable rasterization, largely alleviating the need for vectorized road graph annotations.

\begin{figure}[t]
\centering
\includegraphics[width=0.95\linewidth]{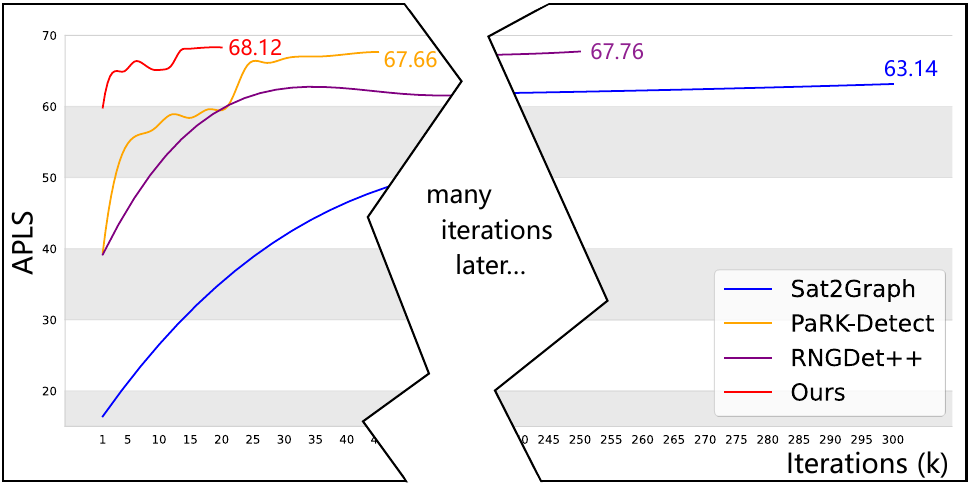}
\caption{Convergence curves on City-Scale dataset.} 
\label{fig_converge_curve}
\end{figure}

In the experiments, we demonstrate that our proposed PaLiS representation clearly set new state-of-the-art performances on two public benchmarks, \ie, the City-Scale~\cite{he2020sat2graph} and SpaceNet~\cite{van2018spacenet}, without paying any extra efforts on the network design. Except for the competitive performance on these two benchmarks, our method only requires 6 GPU hours for the training, significantly reducing the training cost by 32 times for the prior art, RNGDet++~\cite{xu2023rngdet++}. As shown in Fig.~\ref{fig_converge_curve} for the performance evaluation by training iterations on the City-Scale dataset, our proposed method wins after the first $1$K iterations by significant margins and converges to the S.O.T.A. performance after 20K iterations of training.

In summary, our paper made the following contributions:
\begin{itemize}
    \item We propose a novel representation of road graphs, the patched line segment representation, which facilitates the learning of road graphs with the best efficacy in both the training and testing phases.
    \item Based on our patched line segment representation, we present a graph construction strategy for the task of vector road mapping, which takes advantage of the geometric nature of our representation to produce vector graphs without using any additional neural networks for the learning of connectives between keypoints.
    \item Our proposed patched line segment representation is learnable and compatible with the mask-based representation by leveraging a differentiable soft rasterizer, which helps to learn the patched line segments efficiently without introducing additional vector labels.

\end{itemize}

\section{Related Works}
%
\paragraph*{Road Graph Representations.} There have been plenty of studies for vector road mapping, mainly relying on either the rasterized road map or the keypoint/vertex-based graph representations, and derived two categories, the segmentation-based~\cite{mattyus2017deeproadmapper,zhou2018d,mei2021coanet,wang2023ternformer,batra2019improved,cheng2021joint} and the keypoint-based approaches~\cite{he2020sat2graph,he2022td,shit2022relationformer,yang2023topdig,xie2023park}. Regarding the popularity of end-to-end learning for better performance, the state-of-the-art approaches~\cite{he2022td, xu2023rngdet++} mainly learn keypoints (\ie, graph vertices) and the connectivity between vertices while using the rasterized road masks/maps as the additional supervision signals to enhance the feature representation ability of ConvNets. 
Except for the representation ambiguity issue discussed in Sec.~\ref{sec:intro} for prolonged learning schedule, these representations mainly focus on point primitives instead of the line structure of road graphs, thus usually requiring additional design to learn or infer the connectivity between points/pixels. 
Regarding the above issues, we present a novel line-segment-based representation that defines the road graphs in the local image patches while characterizing the structural information of roads using line segments. We show that our well-defined and geometrically-meaningful representation largely facilitates the learning process of vector road mapping with the best efficacy.

\paragraph*{Line Segment Learning and Differentiable Rasterization.} There has been a vast body of literature studying the line segments from both computer vision (CV) and graphics (CG) communities. On one hand, many works study the problem of line segment detection~\cite{xue2019learning,xue2020holistically,XueBWXWZT21,abs-2210-12971}, which is similar to vector road mapping but mainly focuses on the line segment itself instead of the road graphs. On another hand, some CG researchers study the differentiable vector graphics rasterization/rendering~\cite{10.1145/3414685.3417871,Xie0TN14}, in which they aim at using graphic primitives such as points, lines, and curves to represent rasterized digital images. The differentiable rasterization techniques were also applicable to the polygonal shape representation with end-to-end learning in instance segmentation~\cite{lazarow2022instance} and polygonal building extraction~\cite{zorzi2022polyworld}. Our study is inspired by all these studies, but we pay more attention on the well-posedness of the primitive definition for the complicated road graphs/networks. By thinking of local patches, we eventually derive our novel PaLiS representation and set new state-of-the-art performance for the task of vector road mapping.

\section{PaLiS Representation of Road Graphs}

\begin{figure}
    \centering
    \subfigure[The Road Ground Truth]{
    \begin{minipage}[c]{0.42\linewidth}
        \includegraphics[width=1\linewidth]{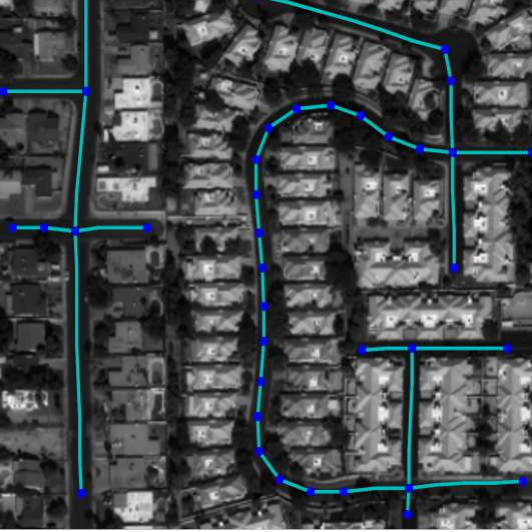}
    \end{minipage}
    \label{fig_patchify_image_gt}
    }
    \hspace{-2mm}
    \begin{minipage}[c]{0.10\linewidth}
        \includegraphics[width=1\linewidth]{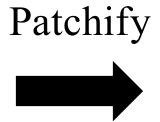}
    \end{minipage}
    \hspace{-2mm}
    \subfigure[\footnotesize PaLiS Representation]{
    \begin{minipage}[c]{0.42\linewidth}
        \includegraphics[width=1\linewidth]{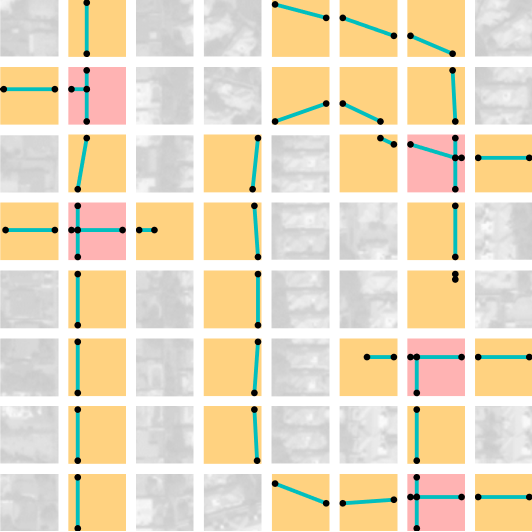}
    \end{minipage}
    \label{fig_construction_overlapping}
    }
    \begin{minipage}[c]{0.8\linewidth}
        \includegraphics[width=1\linewidth]{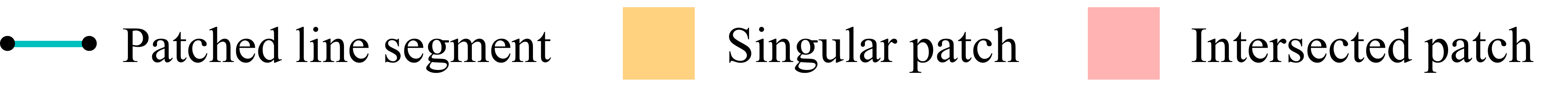}
    \end{minipage}
    \caption{An illustrative figure for the proposed Patched Line Segment (PaLiS) representation. Larger patch size is applied for better illustration.}
    \label{fig_patchify}
\end{figure}

In this section, we elaborate on the proposed PaLiS representation of road graphs. Denoted by the input satellite image $\mathbf{I} \in \mathbb{R}^{3\times H\times W}$ and the corresponding road graph annotation $\mathcal{R} = \{\Gamma_i(t) \in \mathbb{R}^2|t\in [0,1]\}$, where $\Gamma_i(t)$ is a parameterized 2D curve/line, $\Gamma_i(0)$ and $\Gamma_i(1)$ respectively represent the two endpoints of the parameterized curve. We use the local $p\times p$ patches to patch-wisely define the ``key" line segments and eventually form the new PaLiS representation of road graphs. We assume that the patch size $p$ is divisible by $H$ and $W$ without loss of generality.

\subsection{The Main Representation}
By generating a set of $N$ non-overlapping $p\times p$ patches $\{\mathcal{P}_i\}$ where $N = \frac{H}{p}\times\frac{W}{p}$, we define the patched line segment for each local patch $\mathcal{P}_i$. As shown in Fig.~\ref{fig_patchify}(b), there are three cases for each patch $\mathcal{P}_i$ depending on the number of roads passing through the patch, denoted by $\mathcal{N}(\mathcal{P}_i) \in \mathbb{N}$. If $\mathcal{N}(\mathcal{P}_i) = 0$, we term it as the background patch (\ie, the gray patches in Fig.~\ref{fig_patchify}). 
If $\mathcal{N}(\mathcal{P}_i)=1$, we uniquely define its patched line segment, denoted by 
\begin{equation}
    {\rm PaLiS}(\mathcal{P}_i) = (x_i^u,y_i^u,x_i^v,y_i^v) \in \mathbb{R}^4 {\rm if}~ \mathcal{N}(\mathcal{P}_i) = 1.
\end{equation}

\begin{figure}[!b]
    \centering
    \subfigure[$I$-type]{
    \begin{minipage}[b]{0.31\linewidth}
        \includegraphics[width=1\linewidth]{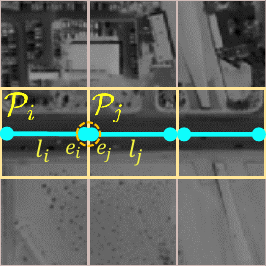}
    \end{minipage}
    \label{fig_construction_type_i}
    }
    \hspace{-2mm}
    \subfigure[$X$-type]{
    \begin{minipage}[b]{0.31\linewidth}
        \includegraphics[width=1\linewidth]{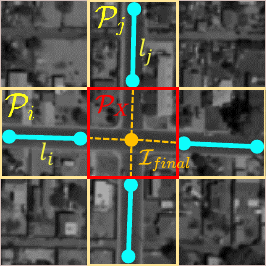}
    \end{minipage}
    \label{fig_construction_type_x}
    }
    \hspace{-2mm}
    \subfigure[$T$-type]{
    \begin{minipage}[b]{0.31\linewidth}
        \includegraphics[width=1\linewidth]{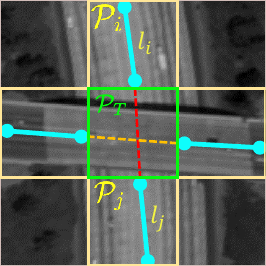}
    \end{minipage}
    \label{fig_construction_type_t}
    }
    \caption{Illustration of different types of foreground patches.  Patched line segments are denoted in cyan markers and connectivities are denoted in dashed markers.}
    \label{fig_graph_construction}
\end{figure}

For those patches that satisfy $\mathcal{N}(\mathcal{P}_i)>1$, we cannot uniquely define their line segments, but we found such patches are playing a key role to construct the expected road graphs. As shown in Fig.~\ref{fig_graph_construction}, we further study the properties of the patches that have $\mathcal{N}(\mathcal{P}_i) \geq 1$. In Fig.~\ref{fig_construction_type_i}, the foreground patches clearly define a (local) straight road without ambiguity. But for the patches that have $\mathcal{N}(\mathcal{P}_i) > 1$, there are two types as shown in Fig.~\ref{fig_construction_type_x} and \ref{fig_construction_type_t}, depending on if there is an annotated ``keypoint" to connect the multiple road paths in one keypoint. If there is such keypoint annotation, we call such a type as the $X$-type. Otherwise, the multiple road paths passing through the patch $\mathcal{P}_i$ will have different elevations like the overpasses, and we called them as the $T$-type patches. 

In summary, the proposed PaLiS representation firstly samples $N$ non-overlapping local patches and identifies the foreground patches by three different types, the $I$-type, $X$-type and $T$-type, and defines the local line segments for the $I$-type patches in the form of $(x_i^u,y_i^u,x_i^v,y_i^v)$ to retain the geometric information of road paths. In the next section, we will show how to learn our proposed PaLiS representation for the task of vector road mapping.

\begin{figure*}[t]
\centering
\includegraphics[width=\linewidth]{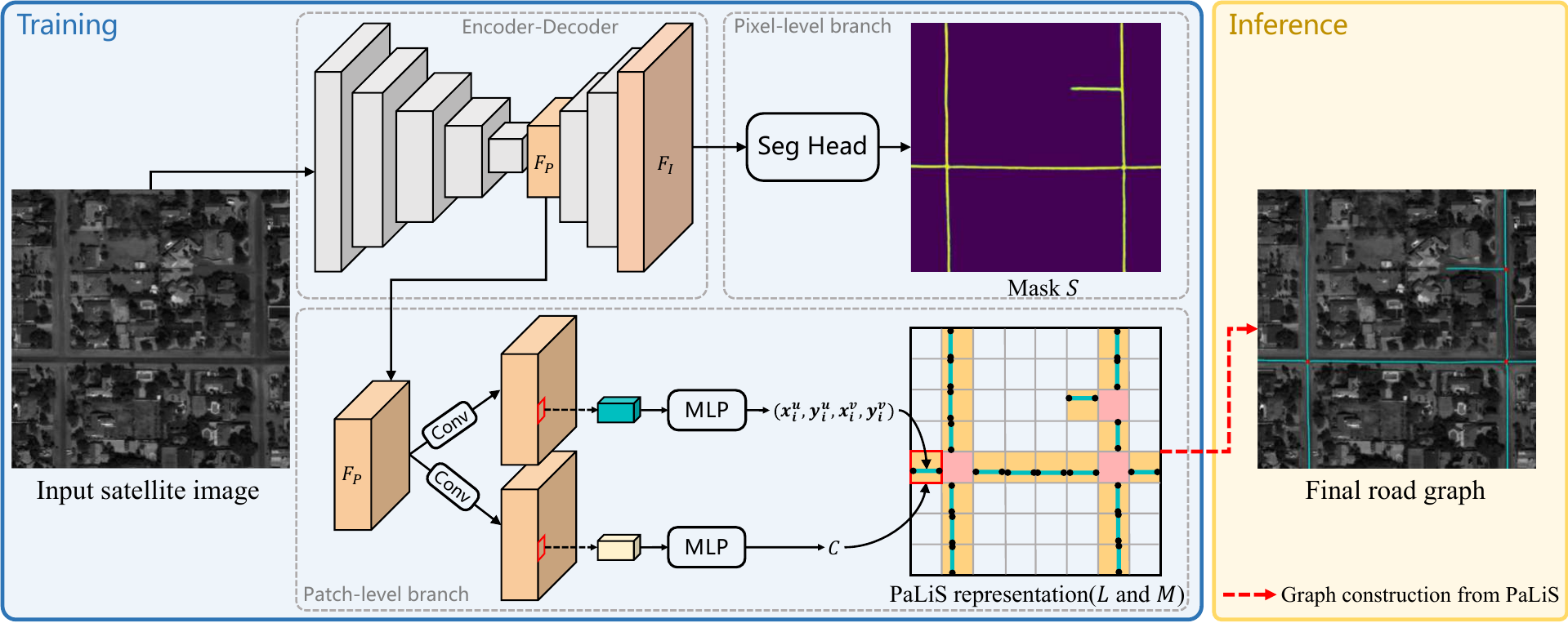}
\caption{Overall pipeline of the proposed method. Given an input image, (1) an Encoder-Decoder network first extracts pixel-level feature maps $\mathbf{F_I}$ and patch-level feature maps $\mathbf{F_P}$. Then, (2) Patch-level branch predicts the line segment and class for each patch with the patch-level features $\mathbf{F_P}$. $(x_i^u,y_i^u,x_i^v,y_i^v)$ and $c$ denote the coordinates of the line segment and type of the patch respectively. (3) Pixel-level branch outputs the binary mask of road centerlines with the pixel-level features $\mathbf{F_I}$. Finally, (4) the road graph is reconstructed from the predicted PaLiS representation. Larger-scale patches are used for better illustration.}
\label{fig_pipeline}
\end{figure*}

\subsection{Road Graph Reconstruction from PaLiS}
Thanks to our geometric PaLiS representation, the road graphs can be reasonably reconstructed without leveraging another subnetwork for the learning of graph connectivity. Here, we hypothesize that the PaLiS representation can be reliably learned and defer the learning details in Sec.~\ref{sec:learning}. 
We developed a geometrically-meaningful scheme to reconstruct the road graphs from our PaLiS representation ({\em see our supp. material for the pseudo code}) by considering the properties of $I$-type, $X$-type and $T$-type foreground patches in the following three cases:
 
\begin{itemize}
    \item As shown in Fig.~\ref{fig_construction_type_i}, we first consider the most common case for the $I$-type patches. For two adjacent $I$-type patches $\mathcal{P}_i$ and $\mathcal{P}_j$, their line segments $\mathbf{l}_i = {\rm PaLiS}(\mathcal{P}_i)$ and $\mathbf{l}_j = {\rm PaLiS}(\mathcal{P}_j)$ are connected with the observation that line segments of adjacent $I$-type patches share a common endpoint. We formulate the rule based on the shape distance ${\rm d_s}(A,B)$, which represents the shortest perpendicular distance between shapes $A$ and $B$. Two line segments are connected if the average of ${\rm d_{s}}(\mathbf{l}_j, \mathbf{e}_i)$ and ${\rm d_{s}}(\mathbf{l}_i, \mathbf{e}_j)$ is less than a given distance threshold $\tau_d$, where $\mathbf{e}_i$ is the endpoint of $\mathbf{l}_i$ close to the line segment $\mathbf{l}_j$ and $\mathbf{e}_j$ is the endpoint of $\mathbf{l}_j$ close to the line segment $\mathbf{l}_i$.
    \item While encountering the $X$-type patch $\mathcal{P}_X$ (\eg, cross roads), line segments surrounding the patch $\mathcal{P}_X$ are extended to an intersection as shown in Fig.~\ref{fig_construction_type_x}. To achieve this, candidate intersections are calculated by pairing up lines segments around the patch $\mathcal{P}_X$. The intersection $\mathcal{I}_{i,j} \in \mathcal{R}^2$ of the line segment pair $(\mathbf{l}_i,\mathbf{l}_j)$ is valid if the two line segments intersect within the patch $\mathcal{P}_X$. And the final intersection $\mathcal{I}_{final}$ is obtained by averaging the position of all candidate intersections and is connected to the surrounding line segments.
    \item Regarding the $T$-type patch $\mathcal{P}_T$ (\eg, overpasses), the layouts with different height are made by the directional and spatial and extension of roads as shown in Fig.~\ref{fig_construction_type_t}. We pair up lines segments around the patch $\mathcal{P}_T$ and the connection of a line segments pair $(\mathbf{l}_i,\mathbf{l}_j)$ is valid if the shape distance ${\rm d_s}(\mathbf{l}_i,\mathbf{l}_j)$ and the angle difference ${\rm d}_{angle}(\mathbf{l}_i,\mathbf{l}_j)$ are less than the distance threshold $\tau_d$ and the angle threshold $\tau_a$ respectively.
\end{itemize}

\section{Learning PaLiS Representations}\label{sec:learning}
%
In this section, we show how to reliably learn the proposed PaLiS representation for vector road mapping in an off-the-shelf ConvNet. We use an encoder-decoder network, DLinkNet~\cite{zhou2018d}, with the lightweight ResNet-34~\cite{he2016deep} as the backbone encoder to extract feature maps for the learning of PaLiS. Fig.~\ref{fig_pipeline} shows the overall pipeline of our approach. For the learning of PaLiS representation, two headnets are respectively leveraged, to classify the patches according to their PaLiS classes, and regress the two endpoints for each $I$-type patch. Apart from the main branches, an auxiliary segmentation head is leveraged to learn the rasterized masks from the final feature maps of the decoder network.

\begin{table*}[t]
\centering
\resizebox{.95\linewidth}{!}{
\begin{tabular}{l|c|c|cccc|cccc}
\toprule
\multirow{3}{*}{Model}  & \multirow{3}{*}{Backbone} & \multirow{3}{*}{Type}  & \multicolumn{4}{c|}{City-Scale}  & \multicolumn{4}{c}{SpaceNet} \\
&                           &                        & \multicolumn{3}{c}{TOPO} & \multirow{2}{*}{APLS $\uparrow$} & \multicolumn{3}{c}{TOPO} & \multirow{2}{*}{APLS $\uparrow$} \\
&                           &                        & P $\uparrow$                    & R $\uparrow$                    & F1 $\uparrow$                   &                                  & P $\uparrow$                    & R $\uparrow$                    & F1 $\uparrow$                   &                                  \\ 
\midrule
DLinkNet~\cite{zhou2018d}                      & ResNet-34  & \multirow{3}{*}{Mask} & 78.63  & 48.07  & 57.42  & 54.08  & 88.42  & 60.06  & 68.80  & 56.93   \\
Orientation~\cite{batra2019improved}           & ResNet-34  &                       & 75.83  & 68.90  & 72.20  & 55.34  & 81.56  & 71.38  & 76.13  & 58.82   \\ 
Seg-DLA~\cite{he2020sat2graph}                 & DLA        &                       & 75.59  & 72.26  & 73.89  & 57.22  & 78.99  & 69.80  & 74.11  & 56.36   \\ \midrule
RoadTracer~\cite{bastani2018roadtracer}        & CNN        & \multirow{6}{*}{Point}& 78.00  & 57.44  & 66.16  & 57.29  & 78.61  & 62.45  & 69.90  & 56.03   \\
Sat2Graph~\cite{he2020sat2graph}               & DLA        &                       & 80.70  & 72.28  & 76.26  & 63.14  & 85.93  & 76.55  & 80.97  & 64.43   \\
TD-Road~\cite{he2022td}                        & ResNet-34  &                       & 81.94  & 71.63  & 76.27  & 65.74  & 84.81  & 77.80  & 81.15  & 65.15   \\
PaRK-Detect~\cite{xie2023park}                 & ResNet-34  &                       & 82.17  & 68.23  & 74.29  & 67.66  & 91.34  & 68.07  & 78.01  & 62.97   \\
RNGDet~\cite{xu2022rngdet}                     & ResNet-50  &                       & 85.97  & 69.78  & 76.87  & 65.75  & 90.91  & 73.25  & 81.13  & 65.61   \\ 
RNGDet++~\cite{xu2023rngdet++}                 & ResNet-101 &                       &85.65  & 72.58  & 78.44  & 67.76  & \textbf{91.34}  & 75.24  & 82.51  & 67.73   \\
\midrule
Ours                                      & ResNet-34 & Line segment & \textbf{86.36}  & \textbf{73.16}  & \textbf{79.08}  & \textbf{68.12}   & 90.05  & \textbf{78.19}  & \textbf{83.70}  & \textbf{69.68} \\
\bottomrule
\end{tabular}
}
\caption{Quantitative results on City-scale Dataset and SpaceNet dataset. Best results are highlighted in bold.}
\label{tbl_main_results}
\end{table*}

\subsection{Identifying Patch Classes/Types}
Our PaLiS representation categorizes the foreground patches into three different types~($I$-type, $X$-type, and $T$-type) for a better understanding of intricate road graph structures. To achieve this, we use a patch classification head, which consists of four convolution layers all with $3 \times 3$ kernels and an MLP layer, to predict the class of each patch. The patch classification head takes patch-level feature maps $\mathbf{F_P}$ as input and produces the patch map $\mathbf{M} \in \mathbb{R}^{C_P \times \frac{H}{p} \times \frac{W}{p}}$, where $C_P$ is the number of patch classes (\ie, $C_P=4$ by considering the background patches). During training, we compute the classification loss by comparing the predicted patch map $\mathbf{M}$ with the corresponding ground truth $\mathbf{M^*}$ which can be easily obtained from the original annotations of the dataset. Cross-entropy loss is employed for $\mathbf{M}$:
\begin{equation}
    \mathcal{L_M} = \rm CE(\mathbf{M}, \mathbf{M^*}).
\end{equation}

\subsection{Line Segments Learning for $I$-type Patches}
With the patch classification head, we focus on the $I$-type patches to learn the patched line segments. It should be noted that although the line segment $\mathbf{l}_i$ for the patch $\mathcal{P}_i$ is in the closed-form for the two endpoints, directly regressing their endpoint coordinates is suboptimal since the data augmentation techniques (like cropping) used in the training phase will incur inefficient computation in terms of cropping the vector road annotations. To avoid this issue, we propose to use the differentiable rasterization techniques to learn the line segment $\mathbf{l}_i$ of the patch $\mathcal{P}_i$ from the mask supervision, similar to ~\cite{lazarow2022instance,zorzi2022polyworld}. It is interesting to see that, although we use the rasterized road mask supervision instead of the vector annotations, such a design is prevailing than the vector annotations. Please move to our ablation studies in Sec.~\ref{sec:exp} for a detailed comparison.

\begin{figure}[t]
\centering
\includegraphics[width=\linewidth]{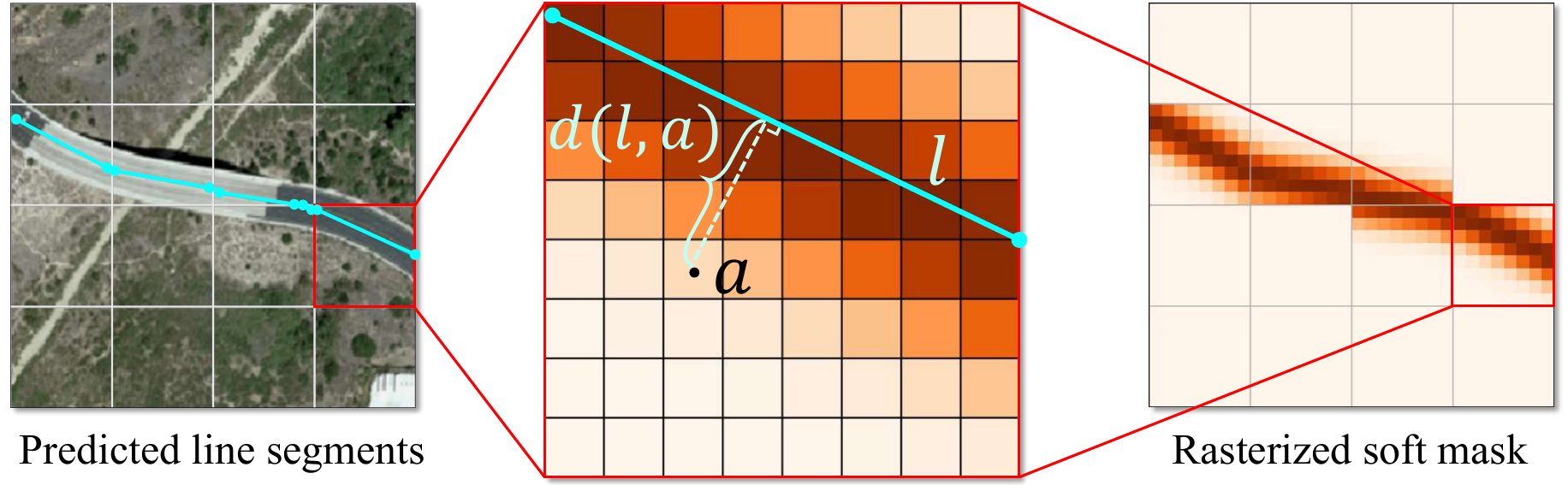}
\caption{Illustration of the rasterization. Darker pixels contribute more to the line segment.}
\label{fig_rasterization}
\end{figure}

By taking the feature map $\mathbf{F}_p$, we set a regression head with four $3 \times 3$ convolution layers and an MLP layer, to predict line segments $\mathbf{L} \in \mathbb{R}^{4 \times \frac{H}{p} \times \frac{W}{p}}$ where 4 is the number of coordinates of line segments. These patched line segments $\mathbf{L}$ are then converted into a soft mask $\mathbf{S_{soft} \in \mathbb{R}^{H \times W}}$ with the proposed rasterizer. As shown in Fig.~\ref{fig_rasterization}, the proposed rasterizer produces a $p\times p$ patch $\mathbf{C}_i \in \mathbb{R}^{p\times p}$, where the scalar value at the pixel $\mathbf{a} = (x,y)$ in the local coordinate of the patch is computed by 
\begin{equation}
    \mathbf{C}_i(\mathbf{a}) = e^{\frac{-{\rm d}^2{(\mathbf{l}_i, \mathbf{a})} \times t}{\tau_{\rm inv}}},
\end{equation}
where ${\rm d}(\mathbf{l}_i,\mathbf{a})$ is the projection distance from the pixel $\mathbf{a}$ to the line segment $\mathbf{l}_i$. $t$ and $\tau_{\rm inv}$ are the projection factor and sharpness factor respectively. We empirically set $t=10$ if the pixel $\mathbf{a}$ is projected outside of the line segment otherwise $t$ is set to 1. The values of projection factor $t$ and the sharpness factor $\tau_{\rm inv}$ are chosen to accurately reflect the position of the line segment in the patch.

The rasterized soft mask $\mathbf{S_{soft} \in \mathbb{R}^{H \times W}}$ is obtained from the contributions of all pixels. During training, we efficiently compute the loss by comparing the soft mask $\mathbf{S_{soft}}$ with the existing ground truth mask $\mathbf{S^*}$ of road centerlines. Similar to BoundaryFormer~\cite{lazarow2022instance}, we employ the DICE~\cite{milletari2016v} loss to measure the difference:
\begin{equation}
    \mathcal{L_L} = {\rm DICE}(\mathbf{S_{soft}}, \mathbf{S^*}).
\end{equation}

The rasterizer and backwards pass are fully implemented in CUDA, ensuring efficiency in the training process.

\subsection{Auxiliary Pixel-level Learning}
In addition to the PaLiS representation, we incorporate the learning of an auxiliary binary mask for road centerlines to extract road information.
We use a segmentation head, which consists of one $3\times 3$ convolution layer and one $1\times 1$ convolution layer followed by a sigmoid function, to predict the binary mask $\mathbf{S} \in \mathbb{R}^{H \times W}$ of road centerlines from the pixel-level feature maps $\mathbf{F_I}$. We compute the loss of the predicted binary mask $\mathbf{S}$ with the ground truth mask $\mathbf{S^*}$ of road centerlines by cross-entropy loss:
\begin{equation}
\begin{split}
    \mathcal{L_S} &= \rm CE(\mathbf{S},\mathbf{S^*})
\end{split}
\end{equation}

The total loss of the PaLiS learning can be summarized as 
\begin{equation}
    \mathcal{L}_{total} = \mathcal{L_S} + \mathcal{L_M} + \mathcal{L_L}.
\end{equation}

\begin{figure*}[t]
    \centering
    \subfigure[Sat2Graph]{
    \begin{minipage}[b]{0.19\linewidth}
        \includegraphics[width=1\linewidth]{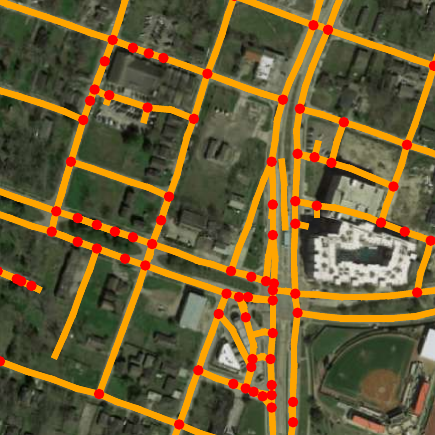}
        \includegraphics[width=1\linewidth]{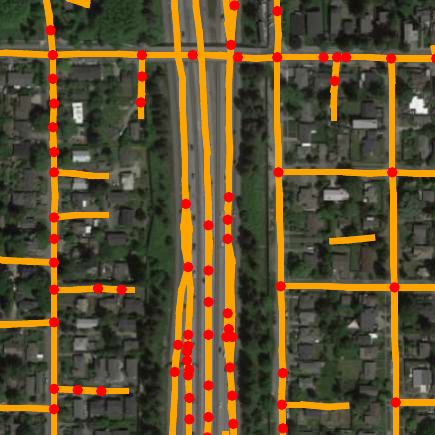}
        \includegraphics[width=1\linewidth]{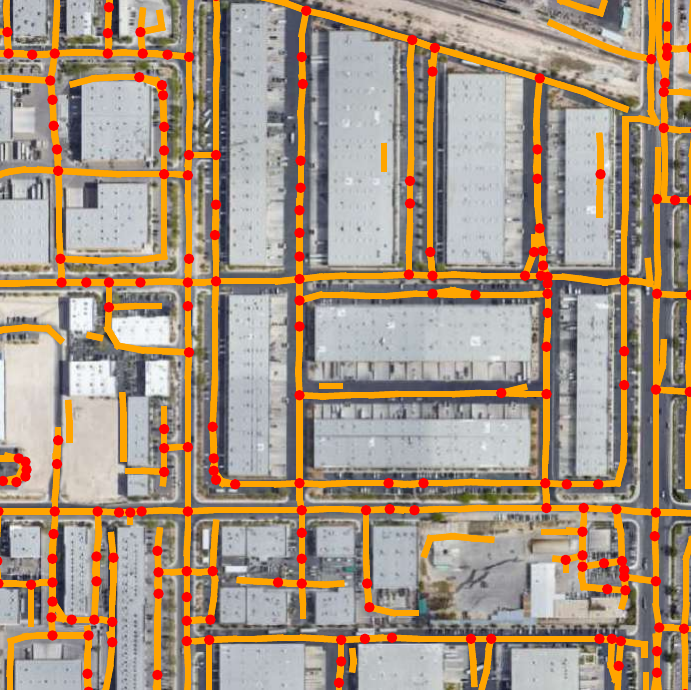}
    \end{minipage}
    }
    \hspace{-2mm}
    \subfigure[RNGDet++]{
    \begin{minipage}[b]{0.19\linewidth}
        \includegraphics[width=1\linewidth]{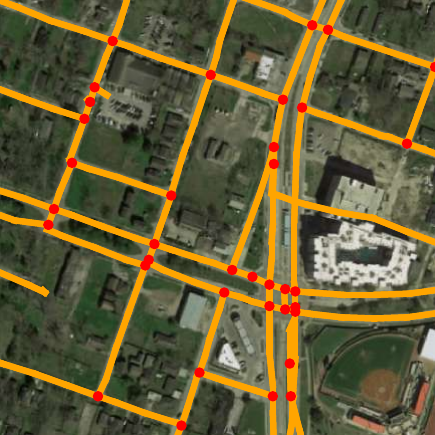}
        \includegraphics[width=1\linewidth]{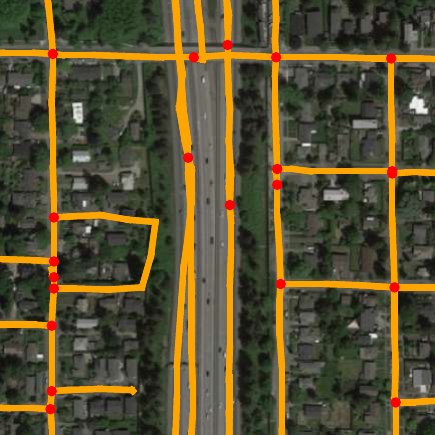}
        \includegraphics[width=1\linewidth]{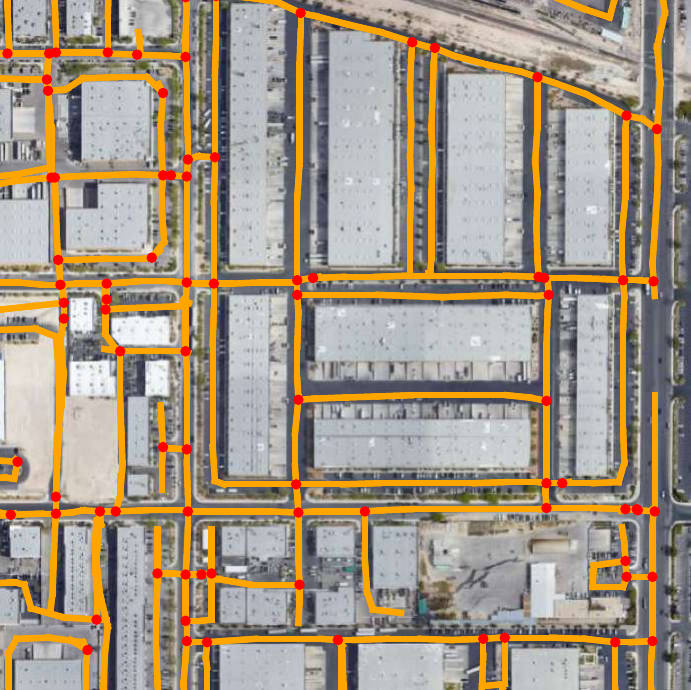}
    \end{minipage}
    }
    \hspace{-2mm}
    \subfigure[PaRK-Detect]{
    \begin{minipage}[b]{0.19\linewidth}
        \includegraphics[width=1\linewidth]{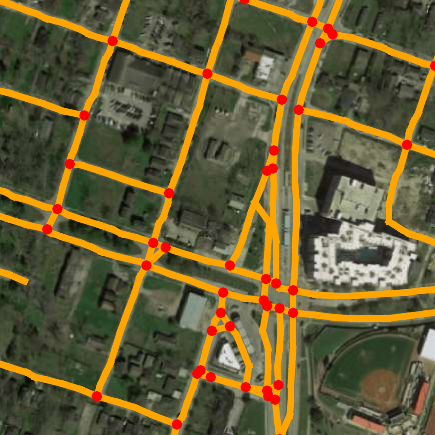}
        \includegraphics[width=1\linewidth]{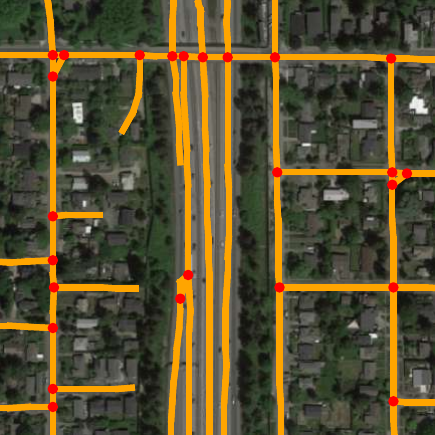}
        \includegraphics[width=1\linewidth]{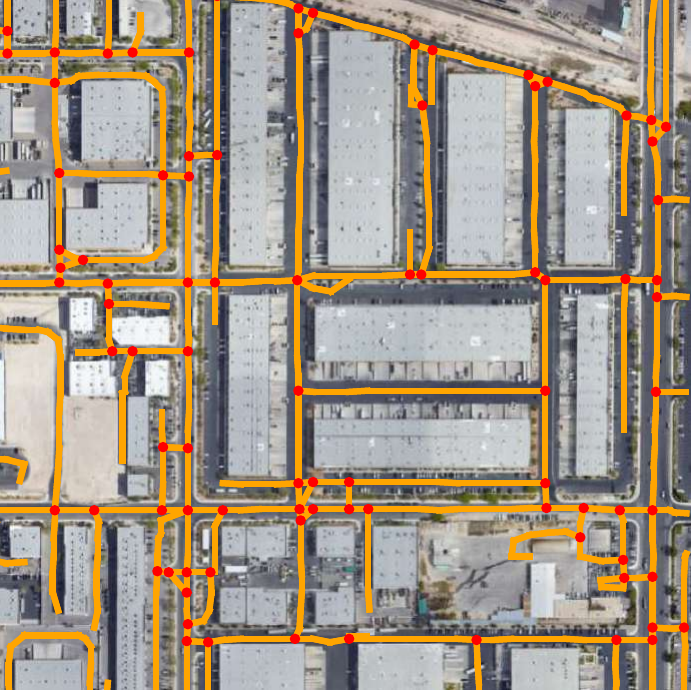}
    \end{minipage}
    }
    \hspace{-2mm}
    \subfigure[Ours]{
    \begin{minipage}[b]{0.19\linewidth}
        \includegraphics[width=1\linewidth]{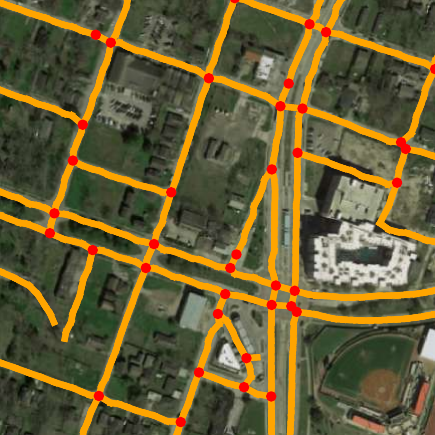}
        \includegraphics[width=1\linewidth]{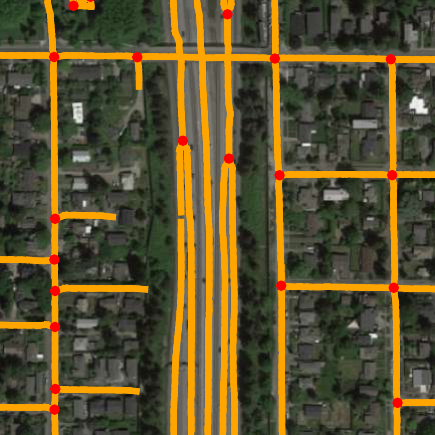}
        \includegraphics[width=1\linewidth]{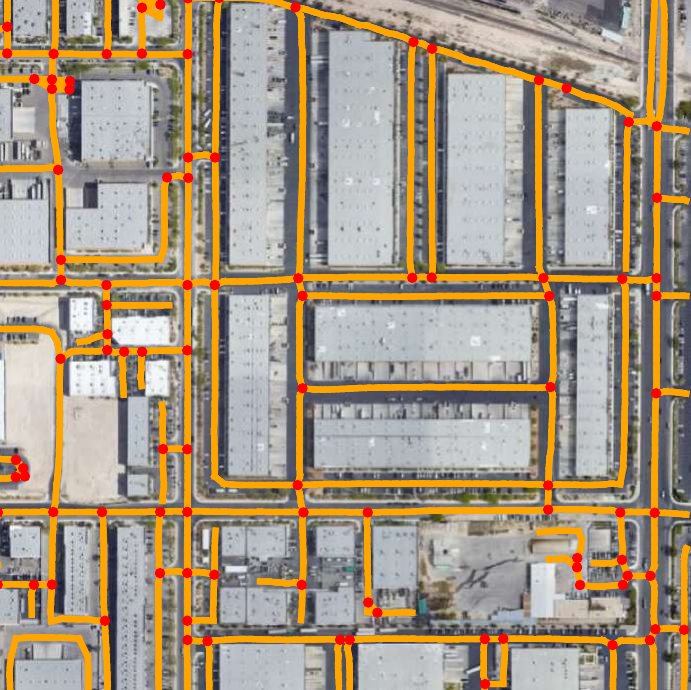}
    \end{minipage}
    }
    \hspace{-2mm}
    \subfigure[Ground Truth]{
    \begin{minipage}[b]{0.19\linewidth}
        \includegraphics[width=1\linewidth]{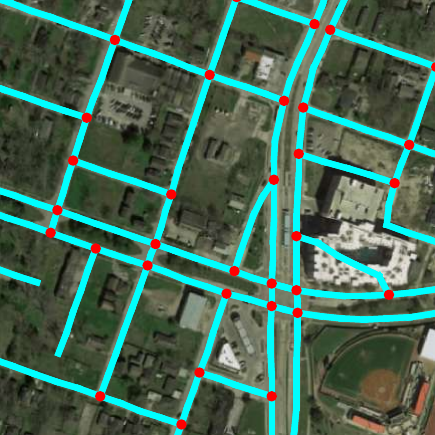}
        \includegraphics[width=1\linewidth]{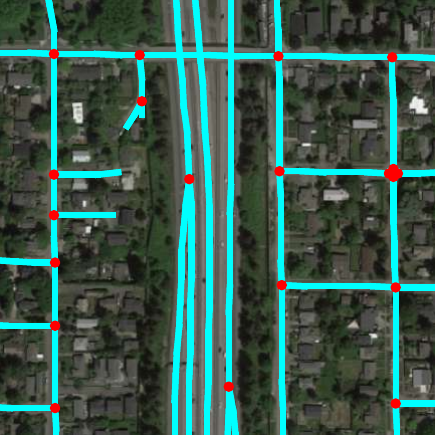}
        \includegraphics[width=1\linewidth]{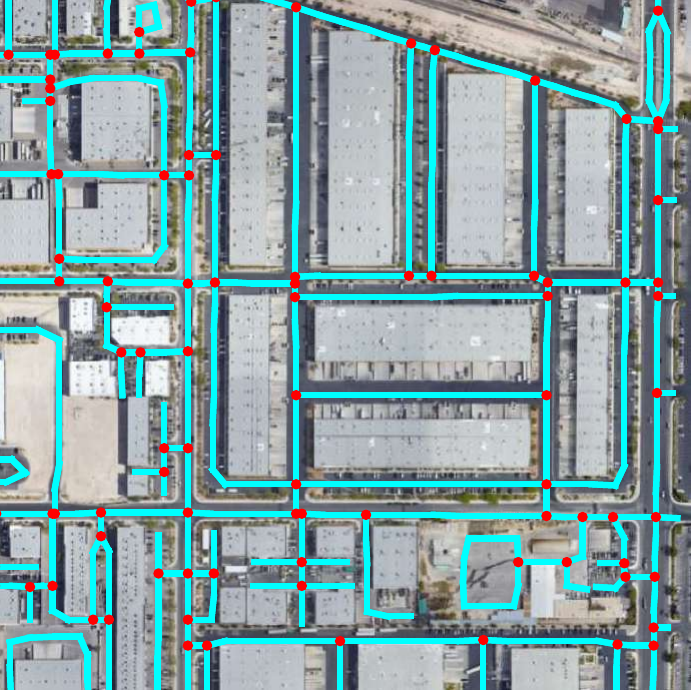}
    \end{minipage}
    }
    \caption{Example of qualitative road network extraction results on City-scale dataset. Predicted road segments are marked in orange lines and intersections are marked in red dots. Our approach leads to reasonable and accurate connected road graphs.}
    \label{fig_cities_compare}
\end{figure*}

\section{Experiments}\label{sec:exp}
%
In this section, we run experiments for our proposed PaLiS-based approach on public benchmarks and provide a comprehensive analysis of our design choices. The implementation details are in our supplementary material.
\subsection{Datasets and Evaluation Metrics}
\paragraph*{Datasets.}
We conduct experiments on two widely used datasets: City-Scale dataset~\cite{he2020sat2graph} and SpaceNet dataset~\cite{van2018spacenet}. City-Scale dataset~\cite{he2020sat2graph} covers $720\ km^2$ area of 20 cities in the United States. It consists of 180 tiles, which we divide into 144, 9, and 27 tiles for training, validation, and testing respectively, following previous methods~\cite{he2020sat2graph,he2022td,xu2023rngdet++}. Each tile of the dataset has the resolution of $2048\times 2048$ pixels, representing 1 meter in the real world. 
SpaceNet dataset~\cite{van2018spacenet} comprises 2549 satellite images, each with the resolution of $400\times 400$ pixels. We use 2040, 127, and 382 images for training, validation, and testing respectively, following the partition used in Sat2Graph~\cite{he2020sat2graph}.

\paragraph*{Evaluation metrics.}
Two quantitative metrics are utilized in the experiments: APLS~\cite{van2018spacenet} and TOPO~\cite{biagioni2012inferring}. 
APLS assesses the overall graph quality by comparing the similarity of shortest paths between two locations on the predicted and ground truth graphs. On the other hand, the TOPO metric (precision, recall, and F1-score) provides a stricter evaluation of detailed topology correctness by measuring the similarity of sub-graphs sampled from a seed location on the ground truth and predicted graphs. Higher scores indicate better performance for both APLS and TOPO metrics.

\subsection{Main Comparisons}
\paragraph*{Quantitative and Qualitative Evaluation.}
We compare our approach to state-of-the-art segmentation- and keypoint-based methods on the City-Scale and SpaceNet datasets. Table~\ref{tbl_main_results} presents the quantitative results. Segmentation-based methods exhibit substantially inferior performance on both TOPO and APLS metrics, because of their heuristic post-processing schema. In contrast, graph-based methods output and refine the graph of road networks directly, gaining better performance on the two metrics. Our method achieves the highest TOPO and APLS scores on the City-Scale dataset, demonstrating superior performance in capturing road network structures with our unified PaLiS representations. Additionally, our approach outperforms all other methods in terms of recall, F1-score, and APLS on SpaceNet dataset, further validating its effectiveness. These consistently superior evaluation results across metrics indicate that our approach generates more precise and complete road graphs both locally and globally. The same conclusions can be drawn from the qualitative comparisons in Fig.~\ref{fig_cities_compare}.

\begin{figure}[t]
    \centering
    \includegraphics[width=\linewidth]{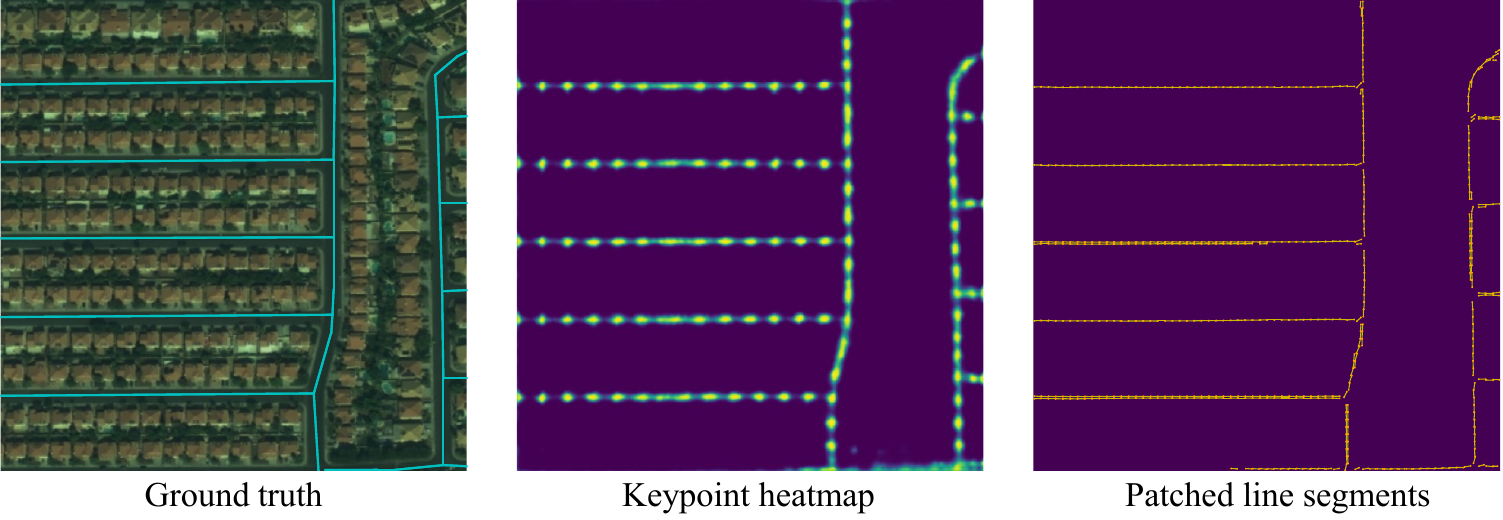}
    \caption{Comparison of keypoints and patched line segments on early stage (10 epoch) of training.}
    \label{fig_element_learning}
\end{figure}

\begin{figure}[t]
    \centering
    \includegraphics[width=\linewidth]{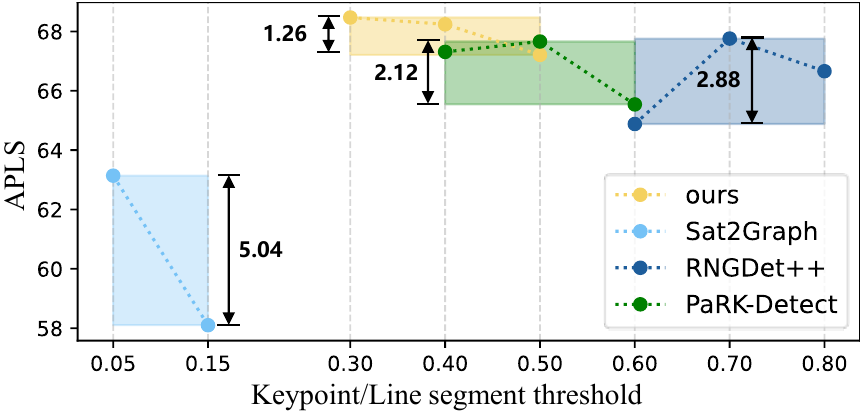}
    \caption{Parameter sensitivity on City-Scale dataset.}
    \label{fig_point_line_threshold}
\end{figure}

\paragraph*{Keypoint-based Graph Representation {\em v.s.} PaLiS.}
Comparisons between keypoints and PaLiS representation during training and testing involved further analysis. Fig.~\ref{fig_element_learning} first visualized the predicted keypoints heatmap and line segments on the early training epoch. Apparently, the learned keypoints heatmap was ambiguous in the early stage of training, whereas the line segments were accurately predicted. Subsequently, we studied the model's sensitivity to thresholds of keypoints (or line segments) prediction by varying the thresholds with the 0.1 step as shown in Fig.~\ref{fig_point_line_threshold}. Notably, our model demonstrated greater stability compared to keypoint-based methods, indicating the robustness of our PaLiS representation during testing.

\paragraph*{Training efficiency.}

The training efficiency is also compared as shown in Fig.~\ref{fig_converge_curve}. The approach relying on our unified PaLiS representation achieves superior performance with considerably fewer training iterations while methods relying on 
keypoints~\cite{he2020sat2graph,xu2023rngdet++,xie2023park} require much more iterations to converge. 

\subsection{Ablation Studies}

\paragraph*{Mask-supervised line segment learning.}
To evaluate the efficacy of the proposed soft rasterizer, we conducted additional experiments using three different types of supervision for line segments learning: unsorted vector labels, sorted vector labels, and mask labels. The unsorted and sorted vector labels are denoted by $(\hat{x}_i^u,\hat{y}_i^u,\hat{x}_i^v,\hat{y}_i^v) \in \mathbb{R}^4$, where the only difference is the direction. Directions of unsorted vector labels are random inherent from origin annotations, while sorted vector labels have consistent directions~($(\hat{x}_i^u,\hat{y}_i^u)$ is always the endpoint on the left). We use L1 loss to compute the difference between the predictions and ground truth vector labels. The results shown in Table~\ref{tbl_supervision} indicate that line segments are learned more precisely with the proposed rasterizer, leading to enhanced connectivity in the graph construction. Furthermore, our approach leverages the existing mask labels to guide the training process of patched line segments, without requiring the generation of vector labels. 

\begin{table}[t]
\centering
\resizebox{.9\linewidth}{!}{
\begin{tabular}{l|cccc}
\toprule
\multirow{2}{*}{Line segments supervision} & \multicolumn{3}{c}{TOPO} & \multirow{2}{*}{APLS $\uparrow$} \\
                            & P $\uparrow$ & R $\uparrow$   & F1 $\uparrow$     &         \\ \midrule
unsorted vector label   & 91.79 & 60.34 & 72.82 & 57.34    \\
sorted vector label     & 91.75 & 66.81 & 77.31 & 65.28    \\
mask label              & 90.05 & 78.19 & 83.70 & 69.68    \\
\bottomrule
\end{tabular}
}
\caption{Comparison results on SpaceNet dataset in association with different supervisions for patched line segments.}
\label{tbl_supervision}
\end{table}

\begin{figure}[!h]
\centering
\includegraphics[width=\linewidth]{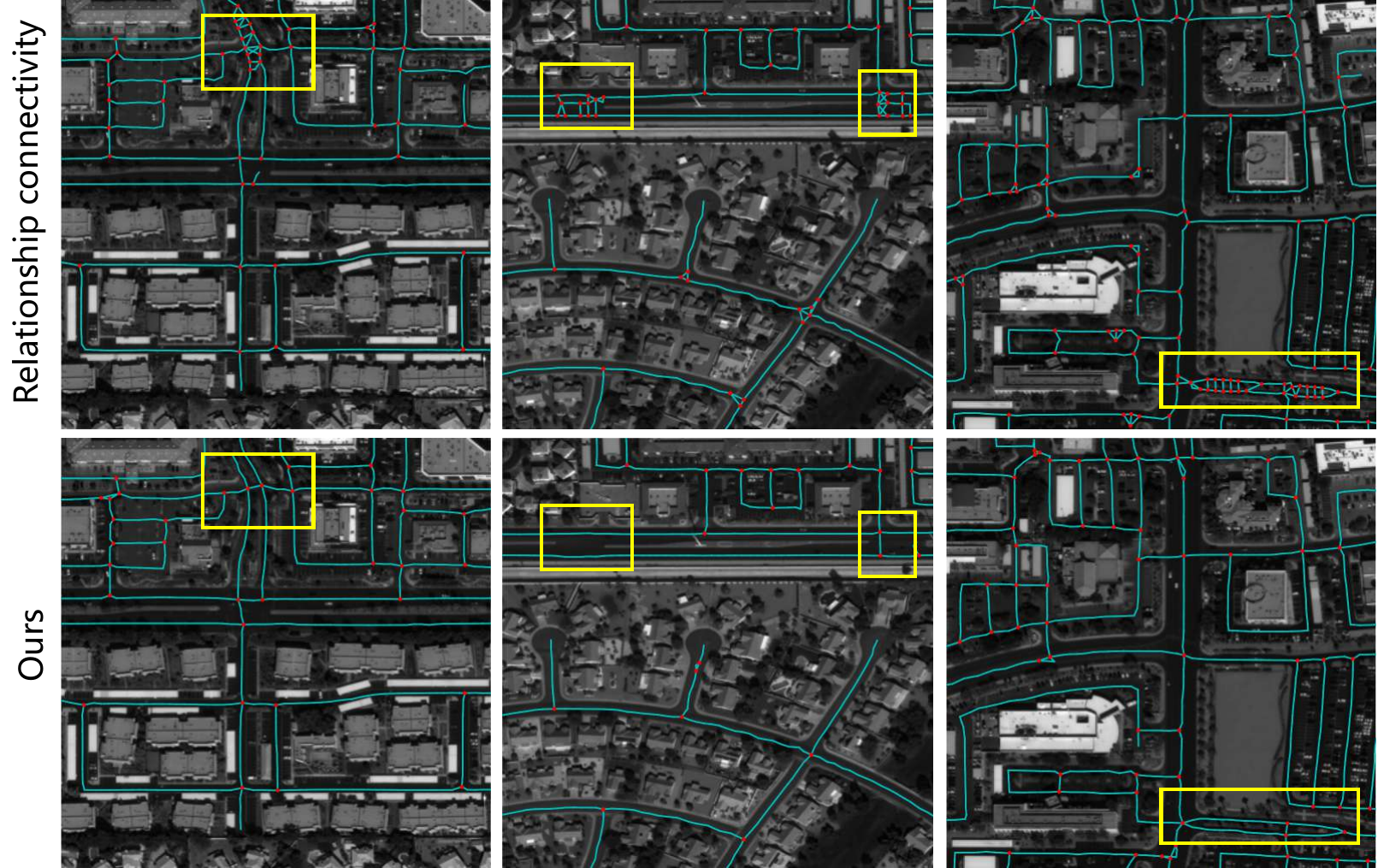}
\caption{Comparison of graph construction strategy.}
\label{fig_graph_contruction_comparison}
\end{figure}

\paragraph*{Graph construction strategy.}
Road graphs can be reconstructed by PaLiS representation~(geometric connectivity) without the learned relationships of patches~(relationship connectivity) used in PaRk-Detect~\cite{xie2023park}. To compare the two different construction strategies, we learned additional relationships of patches following PaRK-Detect~\cite{xie2023park}. The results presented in Table~\ref{tbl_graph_construction} show that our approach outperforms the relationship connectivity on the two metrics, and provides more accurate and reasoned connectivity as shown in Fig.~\ref{fig_graph_contruction_comparison}.

\begin{table}[t]
\centering
\resizebox{.9\linewidth}{!}{
\begin{tabular}{l|cccc}
\toprule
\multirow{2}{*}{\begin{tabular}[c]{@{}c@{}}Graph construction\end{tabular}} & \multicolumn{3}{c}{TOPO} & \multirow{2}{*}{APLS $\uparrow$} \\
                            & P $\uparrow$ & R $\uparrow$   & F1 $\uparrow$     &         \\ 
\midrule
Relationship connectivity      & 88.01 & 79.28 & 83.42 & 68.47    \\
Geometric connectivity~(ours)  & 90.05 & 78.19 & 83.70 & 69.68    \\
\bottomrule
\end{tabular}
}
\caption{Results on SpaceNet of varied connectivity strategy.}
\label{tbl_graph_construction}
\end{table}

\begin{table}
\centering
\resizebox{0.7\linewidth}{!}{
\begin{tabular}{c|cccc}
\toprule
\multirow{2}{*}{Patch Size} & \multicolumn{3}{c}{TOPO} & \multirow{2}{*}{APLS $\uparrow$} \\
                            & P $\uparrow$ & R $\uparrow$   & F1 $\uparrow$     &         \\ \midrule
$4\times 4$                 & 91.88        & 74.23          & 82.12             & 67.25  \\
$8\times 8$                 & 90.05        & 78.19          & 83.70             & 69.68  \\
$16\times 16$               & 82.61        & 77.64          & 80.05             & 67.58  \\
\bottomrule
\end{tabular}}
\caption{Results on Spacenet of varied patch size.}
\label{tbl_patch_size}
\end{table}

\paragraph*{Patch size.}
The patch size serves as a crucial hyper-parameter in our approach. We conducted experiments to assess the impact of patch size, and the results are shown in Table~\ref{tbl_patch_size}. We observe that both smaller and bigger patch sizes cause the inferior performance. This is due to the PaLiS representation with small patch size yields results that are close to mask representation, suffering from the disconnected issue. Whereas PaLiS representation with big patch size struggles to provide precise shape of road graphs. Considering accuracy and efficiency, we set the patch size to 8.

\section{Conclusions}
%
This paper introduces a learning-based approach for vector road mapping using the innovative PaLiS (Patched Line Segment) representation. By leveraging local patches, our approach effectively represents road graphs. Through convolutional neural networks, we achieve state-of-the-art performance on public datasets, with efficient training in just 6 GPU hours. Additionally, the ability of PaLiS representation to learn line segment endpoint coordinates from rasterized road maps suggests a promising direction for large-scale vector road mapping without costly manual annotations in the near future.

{\small 
\bibliography{conference}
}
\appendix
\section*{Supplementary Material}

The supplementary material is organized as follows:
\begin{itemize}
    \item Method Details (Appx.~A): In this section, we provide more information about our method, including pseudo code and additional technical details.
    \item Experimental Results (Appx.~B): Here, we present additional experiments to further illustrate our method and demonstrate the effectiveness of the PaLiS representation.
    \item Visualizations (Appx.~C): This section includes more visual results of comparisons and final graphs constructed by the PaLiS representation.
\end{itemize}

\subsection{A.~More Details}
\label{supp:details}

\paragraph*{Pseudo Code of Graph Reconstruction from PaLiS}
To improve the understanding of the graph reconstruction process from our PaLiS representation, we provide the following pseudo code in Algorithm~\ref{alg_construction}. This algorithm constructs a road graph $G=(V,E)$ based on the predicted patch map $M$ and the patched line segments $L$, where $V$ represents the set of nodes and $E$ represents the set of edges.

\paragraph*{Implementation Details}
We use ResNet-34~\cite{he2016deep} as the backbone and implement our method using PyTorch. During training, input images are resized to $512\times 512$ for both datasets with $8\times 8$ patches. We train the model on single V100 GPU for 6k iterations on SpaceNet and 15k iterations on City-Scale. AdamW~\cite{loshchilov2017decoupled} is applied to optimize the training, where initial learning rate and the weight decay are set to $4\times 10^{-4}$ and $1\times 10^{-4}$ respectively. Data augmentations including random flip and random rotation are adopted following the former works~\cite{he2020sat2graph,xu2023rngdet++}. During inference, the angle threshold $\tau_a$ and the distance threshold $\tau_d$ are set to 15\textdegree and 2 respectively. For the proposed soft rasterizer, the sharpness factor $\tau_{inv}$ is set to 8.

\begin{algorithm}[!t]
    \caption{Reconstruction from PaLiS}
    \label{alg_construction}
    \label{alg_reconstruction}
        \KwIn{Patch map $M$ and patched line segments L}

        $V \gets \emptyset$ \tcp{set of nodes}
        $E \gets \emptyset$ \tcp{set of edges} 

        \For{$\mathcal{P}_i \in M$}{

            \eIf{$\mathcal{P}_i$ is background patch}{
                continue
            }{
                $l_i \gets {\rm PaLiS}(\mathcal{P}_i)$\\
                $N_a \gets$ gather\_neighbor\_I\_patches($\mathcal{P}_i$)\\
            $L_a = \{{\rm PaLiS}(\mathcal{P}_a)|\mathcal{P}_a \in N_a \}$\\
                \If{$\mathcal{P}_i$ is $I$-type patch}{
                    \For{$l_j \in L_a$}{
                        $e_i \gets$ closest endpoint of $l_i$ to $l_j$\\ 
                        $e_j \gets$ closest endpoint of $l_j$ to $l_i$\\
                        \If{${\rm avg}({\rm d_s}(l_i,e_j),{\rm d_s}(l_j,e_i)) \leq \tau_d$}{
                            connect $l_i$ and $l_j$, update $E$
                        }
                    }
                }

                \If{$\mathcal{P}_i$ is $X$-type}{
                $I_{valid} \gets \emptyset$\;
                \For{$(l_m,l_n) \in {\rm pairup}(L_{a})$}{
                    \If{$l_m$ and $l_n$ intersect within $\mathcal{P}_i$}{
                        $I_{m,n} \gets$ intersection of $l_m$ and $l_n$\\
                        $I_{valid} \gets I_{valid} \cup \{I_{m,n}\}$
                    }
                }
                $I_{final} \gets {\rm avg}(I_{valid})$\;
                connect $I_{final}$ and $l_a \in L_{valid}$\\
                update V and E\\
                }

                \If{$\mathcal{P}_i$ is $T$-type}{
                    \For{$(l_m,l_n) \in {\rm pairup}(L_{a})$}{
                        \If{${\rm d_s}(l_m,l_n) \leq \tau_d$ and ${\rm d_a}(l_m,l_n) \leq \tau_a$}{
                            connect $l_m$ and $l_n$, update $E$\\
                        }
                    }
                }   
            }
        }

        Return $G$\\

        \KwOut{Road graph $G=(V,E)$}
\end{algorithm}

\subsection{B.~More Experiments}
\label{supp:exp}

\paragraph*{Auxiliary Pixel-level Branch}
The auxiliary pixel-level segmentation branch, commonly used in previous keypoint-based methods~\cite{he2022td,xu2023rngdet++,xie2023park}, is designed to learning the binary mask of road centerlines to extract more road information. Notably, the binary mask does not directly contribute to the final road graph. To investigate the influence of the segmentation branch within our methodology, we conduct experiments on SpaceNet dataset. The results are shown in Table~\ref{tbl_segmentation_branch}. our approach that excludes the segmentation branch demonstrates inferior performance on both APLS and TOPO metrics. It indicates that the auxiliary pixel-level segmentation branch plays an important role in extracting road information, consequently improving the quality of constructed road graphs.

\begin{table}[h]
\centering
\resizebox{\linewidth}{!}{
\begin{tabular}{c|cccc}
\toprule
\multirow{2}{*}{Segmentation branch} & \multicolumn{3}{c}{TOPO} & \multirow{2}{*}{APLS} \\
        & P      & R      & F1     &                       \\ \midrule
w/o     & 91.28  & 76.45  & 83.21  & 68.77                 \\ 
w/      & 90.05  & 78.19  & 83.70  & 69.68                 \\ \bottomrule
\end{tabular}
}
\caption{Results of ablation study of auxiliary segmentation branch on SpaceNet dataset.}
\label{tbl_segmentation_branch}
\end{table}

\begin{figure}[h]
    \centering
    \includegraphics[width=\linewidth]{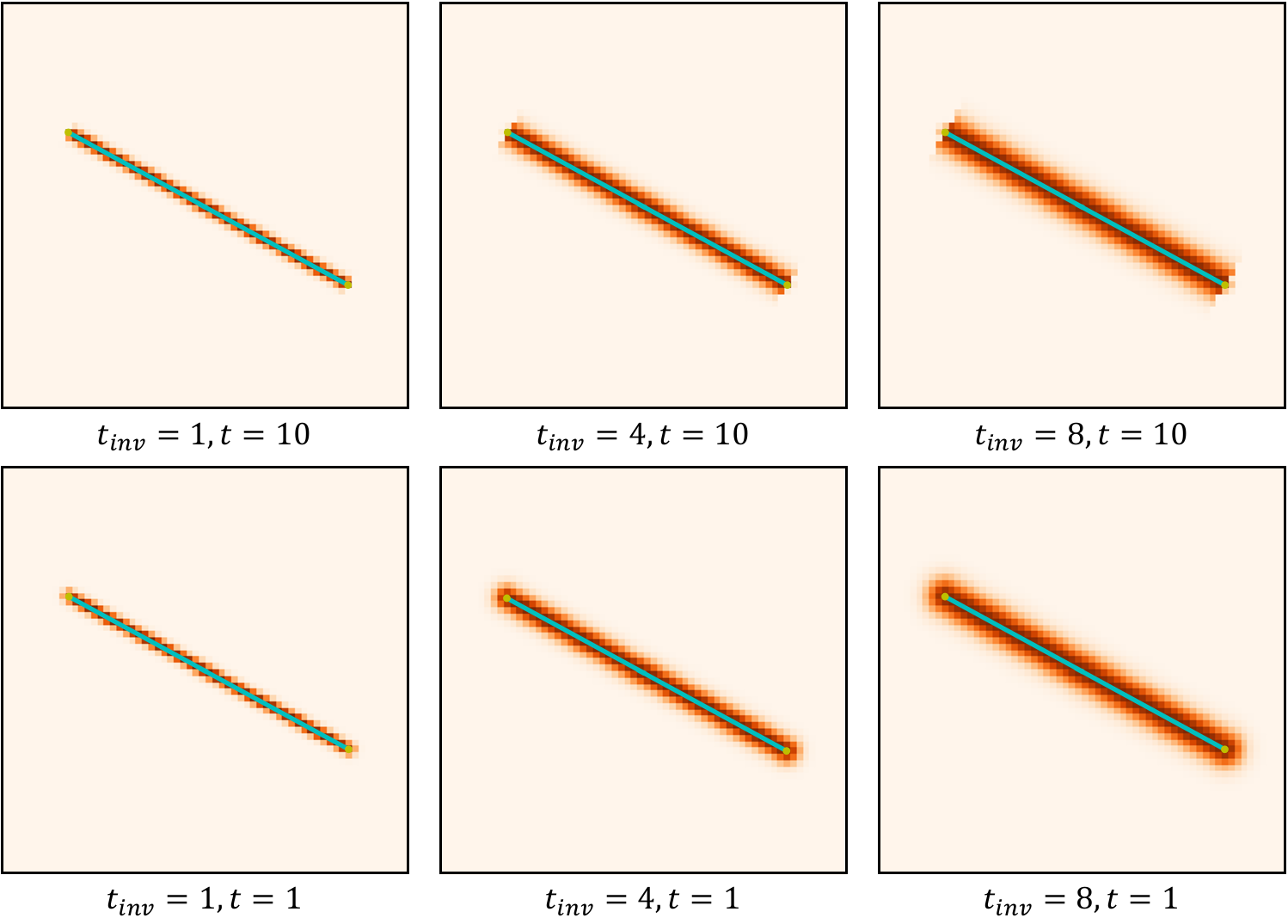}
    \caption{Visualizations of rasterized softmasks with different settings.}
    \label{fig_rasterizer_setting}
\end{figure}

\paragraph*{Effect of Rasterization Settings}
We examined the impact of sharpness ($\tau_{inv}$) and projection ($t$) factors in the rasterization process by conducting additional experiments on the SpaceNet dataset. Table~\ref{tbl_rasterizaion_setting} displays the results. Interestingly, we found that varying the sharpness factor had no noticeable effect on performance for line segment learning or the final graph construction when the projection factor was set to 10. However, our proposed method exhibited a significant decrease in performance without the use of the projection factor. This is due to the fact that the sharpness factor ($t_{inv}$) controls the width of the soft mask from the line segment, while the projection factor ($t$) controls the length of the soft mask from the line segment (as shown in Figure~\ref{fig_rasterizer_setting}). Without the utilization of the projection factor, the rasterized soft mask did not accurately represent the length of the line segment, thus reducing the performance.

\begin{table}[h]
\centering
\begin{tabular}{c|c|cccc}
\toprule
\multirow{2}{*}{$\tau_{inv}$} & \multirow{2}{*}{$t$} & \multicolumn{3}{c}{TOPO} & \multirow{2}{*}{APLS} \\ 
 & & P      & R      & F1     &                       \\ \midrule
0.1      & \multirow{4}{*}{w/}          & 90.45 & 77.77 & 83.63 & 69.30 \\
1        &                              & 90.32 & 78.21 & 83.83 & 69.96 \\
4        &                              & 90.13 & 78.08 & 83.67 & 69.16 \\
8        &                              & 90.18 & 78.13 & 83.73 & 69.68 \\\midrule
8        & w/o                      & 90.57 & 75.17 & 82.15 & 67.10 \\ \bottomrule
\end{tabular}
\caption{Results of different settings for rasterization on SpaceNet dataset.}
\label{tbl_rasterizaion_setting}
\end{table}

\subsection{C.~More Visualizations}
\label{supp:vis}

\begin{figure*}
    \centering
    \includegraphics[width=\linewidth]{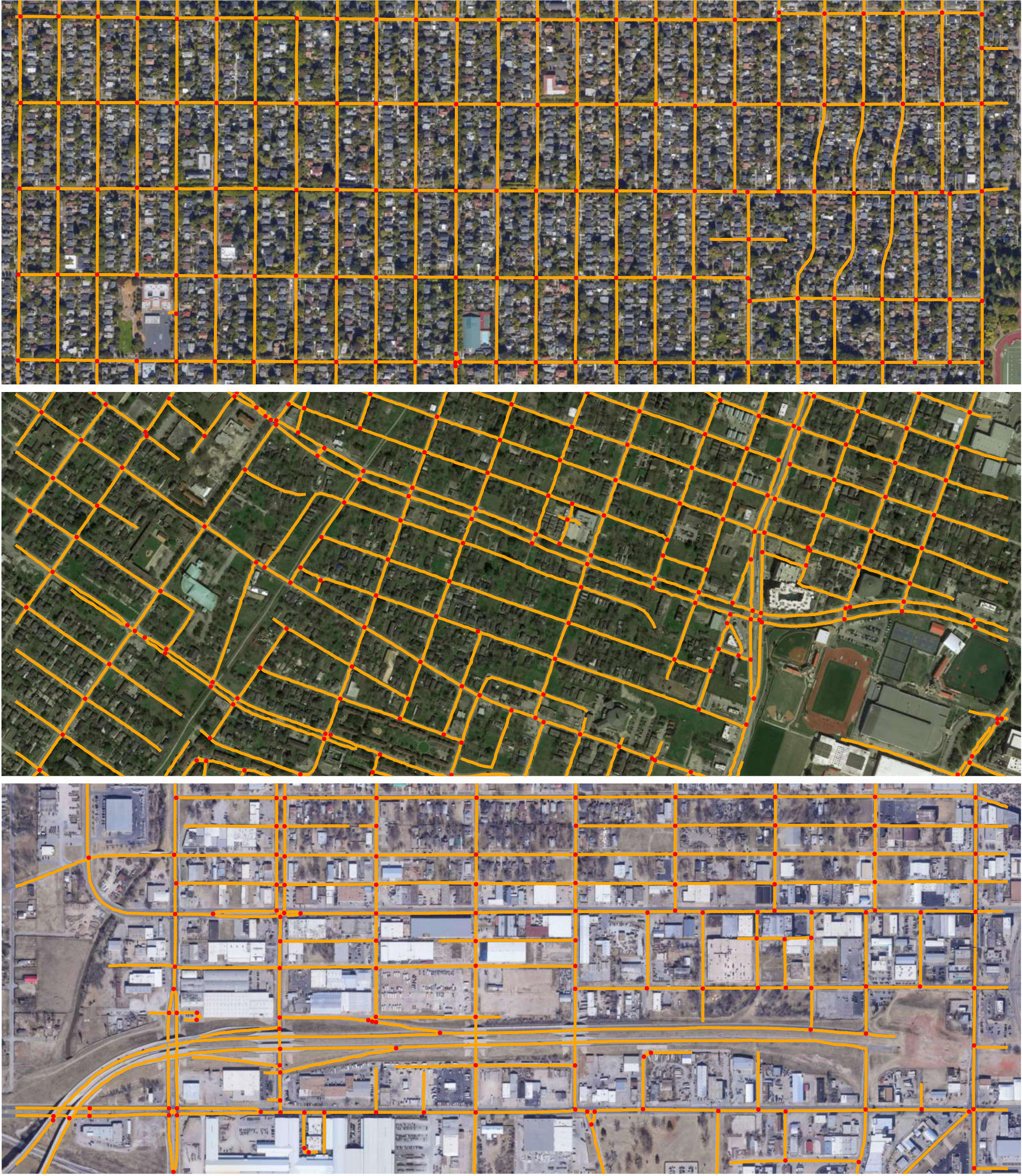}
    \caption{Qualitative results on large-scale images.}
    \label{fig_cities_results1}
\end{figure*}

We present additional qualitative results of the proposed method on the City-Scale dataset. Figures~\ref{fig_cities_results1} showcase the road networks constructed by our method. Our approach excels in accurately predicting complete road networks for large-scale city images. Notably, during the inference phase, our model directly processes the entire $2048 \times 2048$ image as input and outputs a graph with the same resolution, without the need for any crop-and-merge operations. It is worth noting that our model is trained using $512 \times 512$ images, demonstrating its ability to handle large-scale images effectively.

\end{document}